\documentclass[letterpaper]{article} 
\usepackage{aaai23}  
\usepackage{times}  
\usepackage{helvet}  
\usepackage{courier}  
\usepackage[hyphens]{url}  
\usepackage{graphicx} 
\usepackage{natbib}  
\usepackage{caption} 
\usepackage{algorithm}
\usepackage{algorithmic}

\usepackage{amsfonts,amssymb}
\usepackage{multirow}
\usepackage{amsmath}
\usepackage{pifont}

\newcommand{\ie}{\textit{i}.\textit{e}., }
\newcommand{\eg}{\textit{e}.\textit{g}., }

\usepackage{newfloat}
\usepackage{listings}
\DeclareCaptionStyle{ruled}{labelfont=normalfont,labelsep=colon,strut=off} 
\floatstyle{ruled}
\newfloat{listing}{tb}{lst}{}
\floatname{listing}{Listing}
\title{Generalized Semantic Segmentation by Self-Supervised Source Domain Projection and Multi-Level Contrastive Learning}
\author{
    Liwei Yang\textsuperscript{\rm 1}\equalcontrib, 
    Xiang Gu\textsuperscript{\rm 1}\equalcontrib, 
    Jian Sun\textsuperscript{\rm 1,\rm2}\protect\thanks{Corresponding author.}
}
\affiliations{
    \textsuperscript{\rm 1}
School of Mathematics and Statistics, Xi’an Jiaotong University, China\\
\textsuperscript{\rm 2}
Pazhou Laboratory (Huangpu) and Peng Cheng Laboratory, China\\

\{yangliwei, xianggu\}@stu.xjtu.edu.cn, jiansun@xjtu.edu.cn
}

\begin{document}

\maketitle

\begin{abstract}
Deep networks trained on the source domain show degraded performance when tested on unseen target domain data. To enhance the model's generalization ability, most existing domain generalization methods learn domain invariant features by suppressing domain sensitive features. Different from them, we propose a Domain Projection and Contrastive Learning (DPCL) approach for generalized semantic segmentation, which includes two modules: Self-supervised Source Domain Projection (SSDP) and Multi-level Contrastive Learning (MLCL). SSDP aims to reduce domain gap by projecting data to the source domain, while MLCL is a learning scheme to learn discriminative and generalizable features on the projected data. During test time, we first project the target data by SSDP to mitigate domain shift, then generate the segmentation results by the learned segmentation network based on MLCL. At test time, we can update the projected data by minimizing our proposed pixel-to-pixel contrastive loss to obtain better results. Extensive experiments for semantic segmentation demonstrate the favorable generalization capability of our method on benchmark datasets.
\end{abstract}

\section{Introduction}
Deep learning \cite{fcn,deeplab,SETR} has achieved breakthroughs in semantic segmentation, benefiting from the large-scale densely-annotated training images. Nonetheless, obtaining the labeled image data for segmentation is time consuming in real life. For instance, labeling a single image with resolution of $2048\times1024$ in Cityscapes \cite{cityscapes} costs 1.5 hours, and even 3.3 hours for adverse weather conditions \cite{acdc}. An alternative 
solution is training with synthetic data \cite{gtav,synthia}. However, CNNs are sensitive to domain shift and generalize poorly from synthetic to real data.

To deal with this challenge, Domain Adaptation (DA) methods \cite{da3,cycada,fda,da4} align the distributions of source and target domains. However, DA assumes that target data is available in the training process which is hard to fulfill in real-life scenarios. Therefore, Domain Generalization (DG) has been widely studied to overcome this limitation. DG aims to learn a model on source domain data which is generalized well on the unseen target domain. The essence of DG is to learn domain-agnostic features \cite{dg2,dg3,dg4}.
\begin{figure}[t]   
	\centering
	\includegraphics[width=\linewidth,scale=1.0]{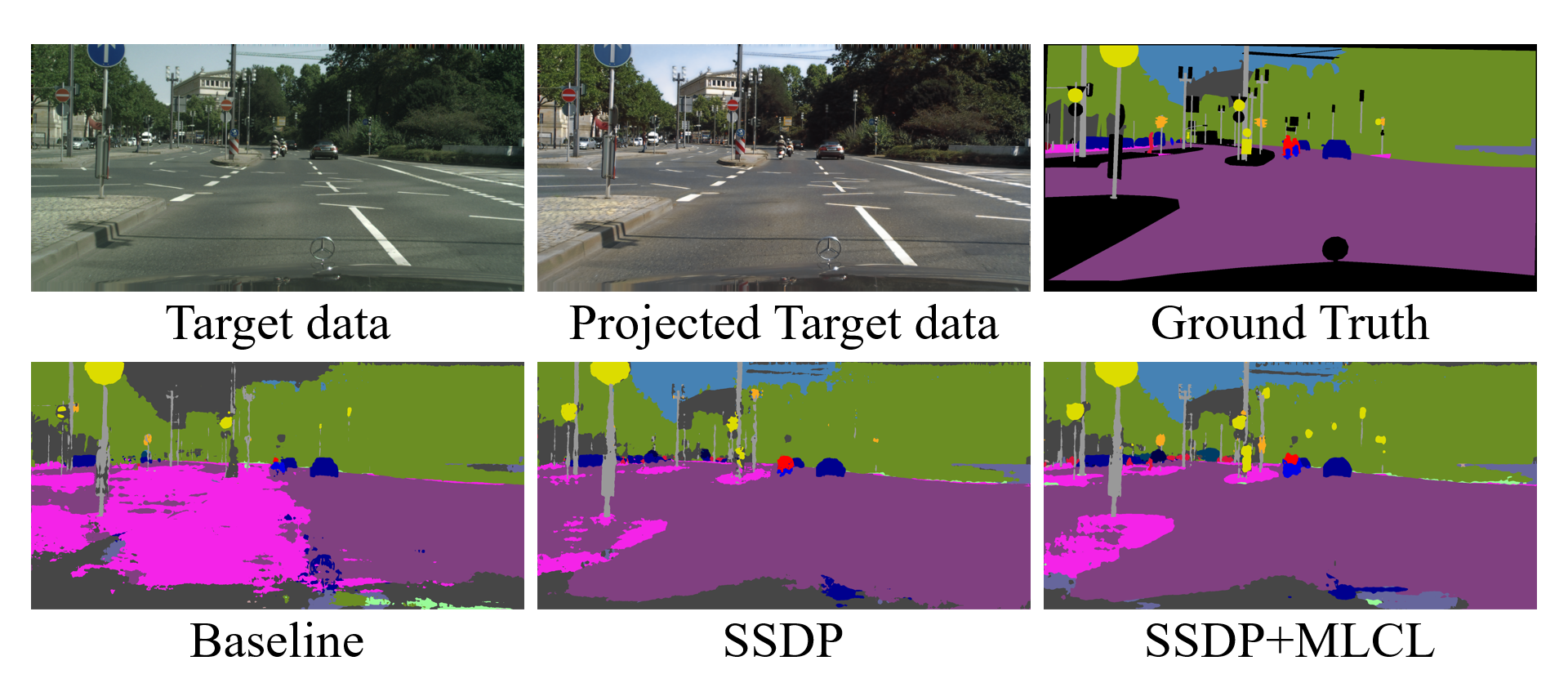}
	\caption{
	Comparison of results for baseline (DeepLabV3+ with backbone ResNet50), our method using Self-supervised Source Domain Projection (SSDP), and using both SSDP and Multi-level Contrastive Learning (MLCL).}
	\label{fig1}
\end{figure}

This work considers Domain Generalization Semantic Segmentation (DGSS), in which we can only use source domain data in training. Existing DGSS methods are mainly divided into three categories. (1) The normalization and whitening-based methods utilize different normalization techniques such as instance normalization or whitening to standardize the feature distribution among different samples~\cite{ibn,isw,dirl,snsw}. (2) Generalizable feature learning methods aim to learn domain-agnostic representation \cite{csg,ptm}. (3) Domain randomization-based methods learn synthetic to real generalization by increasing the variety of training data. \cite{DPRC,gtr,wildnet}. However, domain randomization methods require unlabeled auxiliary datasets for generalization. 

In this paper, we propose Self-supervised Source Domain Projection (SSDP) and Multi-level Contrastive Learning (MLCL) schemes for domain generalization semantic segmentation. Specifically, we first design SSDP, aiming to learn a projection to map the unseen target domain data to the source domain by projecting augmented data to its original data in the training phase. Secondly, for augmented data projected to the source domain, we further propose MLCL to learn a better generalizable segmentation network by contrasting the features with the guidance of labels at the pixel, instance and class levels.  At test time, given an unlabeled target domain image, we first project it onto the source domain by SSDP, then segment it by the learned semantic segmentation model. Extensive experiments show that our SSDP and MLCL schemes improve the generalization performance of our segmentation model.  Figure~\ref{fig1} illustrates an example of segmentation results by the baseline method and its improved versions respectively using SSDP and SSDP+MLCL. Our code is available at https://github.com/liweiyangv/DPCL.

The main contributions can be summarized as follows.
\begin{itemize}
\item We propose a Self-supervised Source Domain Projection (SSDP) approach for projecting data onto the source domain, to mitigate domain shift in the test phase.
\item We propose a Multi-level Contrastive Learning (MLCL) scheme, which considers the relationship among pixels features, instance prototypes and class prototypes. In particularly, we propose to deal with pixel-to-pixel contrastive learning as a Transition Probability Matrix matching problem.
\item We apply our method to urban-scene segmentation task. Extensive experiments show the effectiveness of our DPCL for domain generalization.

\end{itemize}
\section{Related Works}

\subsection{Domain Generalization}
Domain generalization attempts to improve model generalization ability on the unseen target domain. As for classification task, domain generalization is mainly based on domain alignment of source domains to learn domain-invariant features \cite{dgadverarialbased1,dgadverarialbased2}, meta-learning to learn generalizable features \cite{dg3,dg4}, or data augmentation to expand source data to improve generalization capabilities \cite{pden}.

As for semantic segmentation, the existing domain generalization methods can be classified into three categories: 1) Normalization and whitening based-methods utilize Instance Normalization (IN) or Instance Whitening (IW) to standardize global features by erasing the style-specific information and prevent model overfitting on the source domain \cite{ibn,isw,dirl,snsw}. For instance, 
ISW \cite{isw} utilizes IW to disentangle features into domain-invariant and domain-specific parts, and normalize domain-specific features. DIRL \cite{dirl} proposes a sensitivity-aware prior module to guide the feature recalibration and feature whitening, and learns style insensitive features. \cite{snsw} designs a semantic normalization and whitening scheme to align category-level features from different data. 2) Generalizable feature learning-based methods focus on learning domain-invariant features, such as utilizing attention mechanism \cite{csg} or meta-learning framework \cite{ptm}. 
3) Randomization-based methods synthesize images with different styles to expand source domain \cite{DPRC,gtr,wildnet}. DPRC \cite{DPRC} randomizes the synthetic images with the styles of real images and learns a generalizable model. 
WildNet \cite{wildnet} leverages various contents and styles from the wild to learn generalized features. Different from domain randomization methods, which utilize auxiliary dataset to expand source domain data, we adopt a self-supervised scheme to train a source domain projection network, which projects data with different distributions onto the source domain. Based on the projected data, we further propose a multi-level contrastive learning strategy to learn discriminative features.

\subsection{Test-Time Adaptation} Test-time adaptation aims to improve the performance of source trained model against domain shift with a test-time adaptation strategy. Existing methods are mainly designed for classification task and can be categorized in two ways. (1) Update model's parameters at test time by utilizing self-supervised loss. Tent \cite{tent} adopts entropy minimization to fine-tune BN layers in the test phase. Adacontrast \cite{adacontrast} conducts test-time contrastive learning and learns a target memory queue to denoise pseudo label. (2) Learn the model to adapt to test data without using extra loss at test time. For example, \cite{singletest} learns to adapt the model's parameters based on only one test data using the meta-learning framework. Different from the above adaptation schemes, we update the target data by projecting it onto the source domain by our SSDP network, then we iterate the projected data by our pixel-to-pixel contrastive loss, while fixing the parameters of learned models.

\begin{figure*}[htbp]   
	\centering
	\includegraphics[width=0.93\linewidth,scale=1.00]{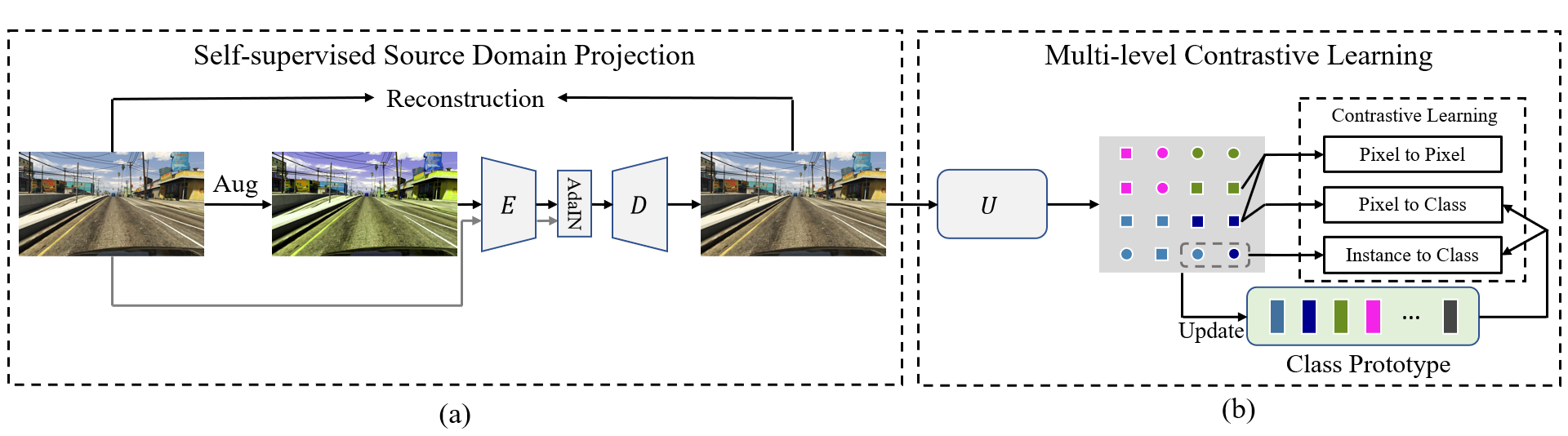}
	\caption{Overview of our proposed DPCL. (a) Self-supervised Source Domain Projection (SSDP) sub-network aims to project data onto the source domain. $E, D$ are the encoder and decoder of SSDP. (b) Multi-level Contrastive Learning (MLCL) based on projected data attempts  to learn discriminative features. $U$ is the feature extractor of segmentation network.}
	\label{framework}
\end{figure*}
\subsection{Contrastive Learning} Contrastive learning has shown compelling performance in representation learning \cite{instdist,simclr,simclrv2,mocov2}. Supervised contrastive learning \cite{scl} pulls the sample pairs 
in the same class closer and pushing away the negative pairs which have different labels to learn discriminative features.~\cite{sclseg,category_aware_con_seg} utilizes supervised contrastive learning scheme in semantic segmentation to constrain pixel-level features. Except for pixel-wise contrastive learning, recent works also utilize other contrastive learning for segmentation, such as prototype-wise \cite{region_aware_con_seg,dual_prototype_con_seg} or distribution-wise \cite{ distribution-aware_con_seg}. Besides traditional InfoNCE loss, \cite{augmix,js_noisy_labels} propose to minimize Jensen-Shannon (JS) divergence among the predictive distributions of samples with different augmentation strategies to learn a robust model. Different from recent work, we define multi-level contrastive learning for pixel features, instance prototypes and class prototypes. Specifically, we reformulate pixel-to-pixel contrastive learning based on transition probability matrix, which shows a better generalization ability in the experiments. 

\section{Method}

In this paper, we focus on a Single-source Domain Generalization (SDG) setting. We denote our source domain as $\mathcal{S}$ and unseen target domain as $\mathcal{T}$. 
Notably, $\mathcal{T}$ has different distribution with the source domain. $\mathcal{S}$ can be represented as $\{(x_{i}, y_{i})\}^{n}_{i=1}$, where $(x_{i}, y_{i})$ denote the $i$-th image and its pixel-wise class label, $n$ is the number of samples in $\mathcal{S}$. SDG aims to train the segmentation model on $\mathcal{S}$ and generalize it to the unseen target domain $\mathcal{T}$.

The proposed method DPCL mainly has two components. As shown in Fig.~\ref{framework}, we first utilize a Self-supervised Source Domain Projection (SSDP) block to project data from other distributions to the source domain. Then, we propose a Multi-level Contrastive Learning (MLCL) scheme to learn discriminative features based on projected data. Next, we will explain our formulation and each module in detail.

\subsection{Self-supervised Source Domain Projection}
The SSDP aims to project data onto the source domain to mitigate domain shift at test time. Since target data is not available in training, we can not directly obtain a style transfer network from target to source like domain adaptation methods \cite{cycada}. In this paper, we adopt a data augmentation strategy to generate data with different distributions from the source domain, and project augmented data to its corresponding original data in the source domain. 

\begin{figure}[t]   
	\centering
	\includegraphics[width=\linewidth,scale=1.0]{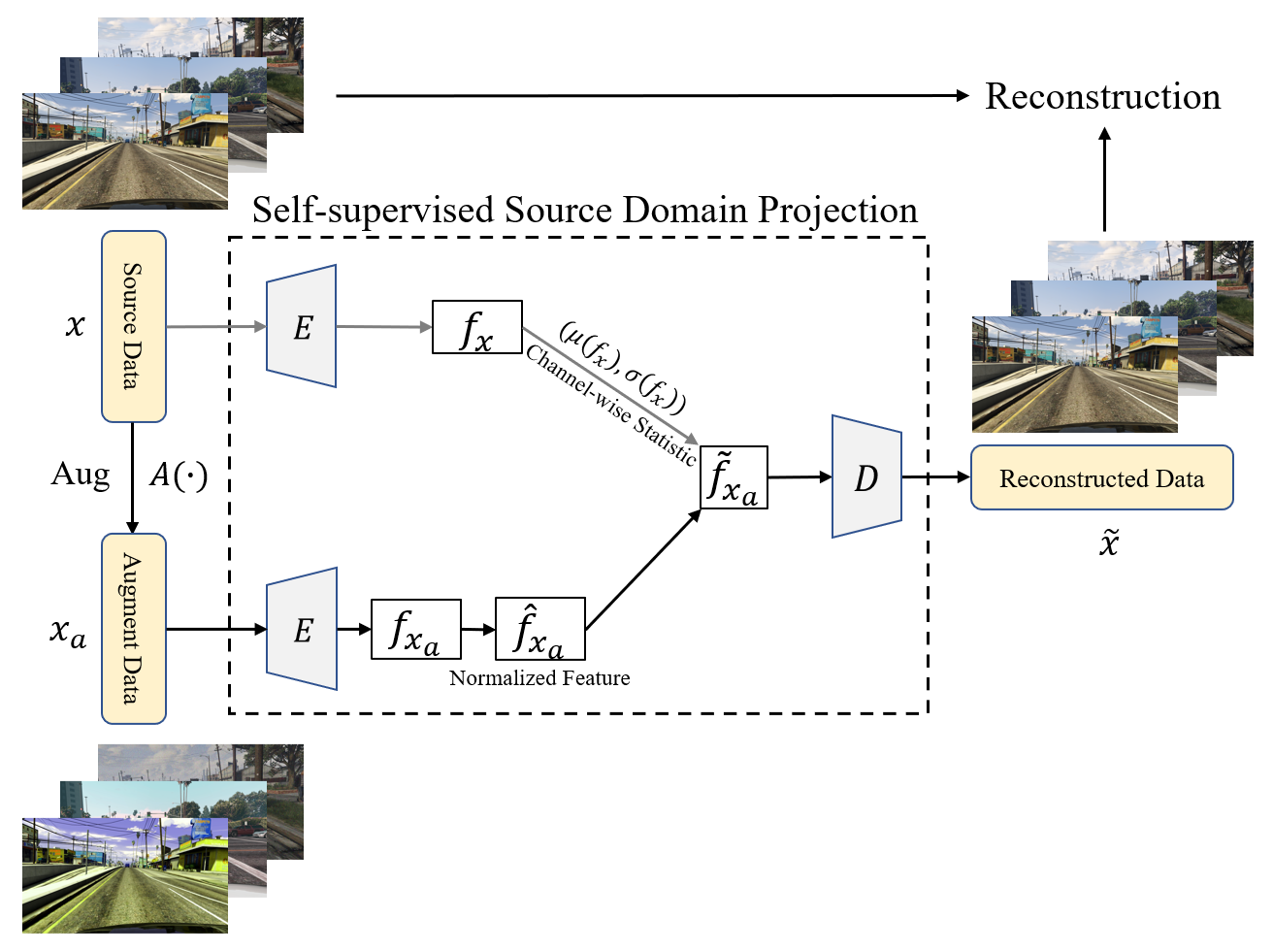}
	\caption{Illustration of Self-supervised Source Domain Projection sub-network. $f_x, f_{x_a}$ are the features of source data $x$ and augmented data $x_a$, $\hat{f}_{x_a}$ is the feature after instance normalization of $f_{x_a}$. $\mu(f_x),\sigma(f_x)$ are the channel wise mean and standard deviation of $f_x$. $\tilde{f}_{x_a}$ is the renormalized feature of $\hat{f}_{x_a}$. $\tilde{x}$ is the reconstructed original image.}
	\label{ae_pic}
\end{figure}
We denote our SSDP as a mapping $F:\mathcal{T}\rightarrow\mathcal{S}$. Given a target data $x$, it aims to make $F(x)$ close to the source domain $\mathcal{S}$. Since target domain data is unavailable in training, we use data augmentation over source domain data to simulate domain shift in the training phase. We project the augmented data to original data to learn our SSDP. Specifically, our design of SSDP is shown in Fig.~\ref{ae_pic}. We denote  $x_a=A(x)$ as the augmented data, where $A$ is an augmentation function. We input both original data $x$ and augmented data $x_a$ into encoder $E$ of SSDP and get feature $f_x$ and $f_{x_a}$. As for $f_{x_a}$, we adopt instance normalization to get normalized feature $\hat{f}_{x_a}$ to eliminate its style information. Meanwhile, we calculate channel-wise standard deviation and mean of feature $f_x$ which contain style information of $x$ as affine parameters to transform normalized feature $\hat{f}_{x_a}$. We assume the transformed feature $\tilde{f}_{x_a}$ contains content information of $x_a$ and style information of $x$. Then we input $\tilde{f}_{x_a}$ into decoder $D$ to get reconstructed original data $\tilde{x}$. Since we only utilize data augmentation to create sample pair $x$ and $x_a$, our scheme of training SSDP can be regarded as a self-supervised way.

Formally, we use a standard instance normalization to get the normalized feature $\hat{f}_{x_a}$ by
\begin{equation}
    \hat{f}_{x_a}=\frac{f_{x_a}-\mu(f_{x_a})}{\sigma(f_{x_a})}
\end{equation}
where $\mu(f_{x_a}),\sigma(f_{x_a})$ are the channel-wise mean and standard deviation of feature $f_{x_a}$, then we use the same statistics of $f_x$ to transform normalized feature $\hat{f}_{x_a}$ by
\begin{equation}\label{adain}
    \tilde{f}_{x_a} = \sigma(f_x)\hat{f}_{x_a}+\mu(f_{x})
\end{equation}
Then we input the transformed feature $\tilde{f}_{x_a}$ into decoder $D$ and get reconstructed data $\tilde{x}$. In the experiment,we use $L_1$ loss for enforcing image reconstruction:
\begin{equation}\label{loss1}
\mathcal{L}_{recon}=||x-\tilde{x}||_1
\end{equation} 
Different from the traditional autoencoder, the input of our SSDP is the augmented data and original data, the output is the original data in the source domain. We adopt AdaIN \cite{adain} in the feature space to make SSDP project augmented data to the original data.

In the test phase, we do not have the paired source data $x$ to provide source style information for each target data $x_t$. We use mean and standard deviation cluster center of source data features to alter $\sigma(f_x)$ and $\mu(f_{x})$ in Eq.~\eqref{adain}. Specifically, we cluster the mean and standard deviation of training data features into $q$ centers after training over the source domain. We denote mean cluster centers as $\mu_{\mathcal{S}}=\{\mu_1,\mu_2,...,\mu_q\}$, standard deviation centers as $\sigma_{\mathcal{S}}=\{\sigma_1,\sigma_2,...,\sigma_q\}$. Given a target data $x_t$, we use $L_2$ distance to find the closest center $\hat{\mu}$ of $\mu(f_{x_t})$ in $\mu_{\mathcal{S}}$, \ie
\begin{equation}
    \hat{\mu}=\mathop{\arg\min}\limits_{\tilde{\mu}}||\tilde{\mu}-\mu(f_{x_t})||_{2}, \tilde{\mu}\in \mu_{\mathcal{S}}
\end{equation}
We can get the closest standard deviation center $\hat{\sigma}$ in the same way. Then we use $\hat{\mu},\hat{\sigma}$ to transform the normalized feature $\hat{f}_{x_t}$ by using Eq.~\eqref{adain}, and get the projected data by sending the renormalized feature $\tilde{f}_{x_t}$ into decoder $D$.


\subsection{Multi-level Contrastive Learning}

Based on the projected data by SSDP, we further propose a multi-level contrastive learning scheme for learning discriminative features.  Using traditional cross-entropy as task loss only penalizes pixel-wise predictions independently but ignores semantic relationships among pixels. To investigate the semantics at different levels and their relations, we propose  multi-level contrastive learning for learning model of semantic segmentation in the feature space. Our segmentation model consists of feature extractor $U$ and classifier $H$. 

Different from image classification, semantic segmentation aims to predict class label for each pixel, and there may exist more than one instance in an image to be segmented. We consider the semantic class relationship among multi-level features, including pixel, instance and class levels to learn discriminative and generalizable features. Specifically, we adopt prototype for instance-level and class-level feature representations by average pooling features in each connected region or total area of each class in each image according to the ground truth segmentation mask. 

\textbf{Construction of Class Prototype.}
Taking class-level prototype as example, we calculate prototype by average pooling the features in each class region:
\begin{equation}\label{equ4}
p^k=\frac{\sum_{i=1}^{H^{\prime}W^{\prime}}y^k_{z_i}z_i}{\sum_{i=1}^{H^{\prime}W^{\prime}}y^k_{z_i}}
\end{equation}
where $H^{\prime}, W^{\prime}$ respectively denotes height and width of feature map. $y_{z_i}$ is one-hot label for pixel feature $z_i$, \ie $y^k_{z_i}=1$, when $z_i$ belongs to class $k$. To obtain the class prototype in the whole training dataset, we update class prototypes using moving average strategy by
$$\hat{P}^k=\gamma P^k+(1-\gamma)p^k$$
where $\hat{P}^k$, $P^k$ are the updated and historical class prototype for class $k$, $p^k$ is $k$-th class prototype calculated in the current training batch, $\gamma$ is momentum set as 0.999. 

We next present our multi-level contrastive learning loss considering pixel-to-pixel, pixel-to-class and instance-to-class feature relations in the feature maps. In the following paragraphs, the features and prototypes are $l_2$ normalized. The ``pixel'' in this work represents the pixel in the feature maps instead of the original image grid. 
\begin{figure}[t]   
	\centering
\includegraphics[width=\linewidth,scale=1.0]{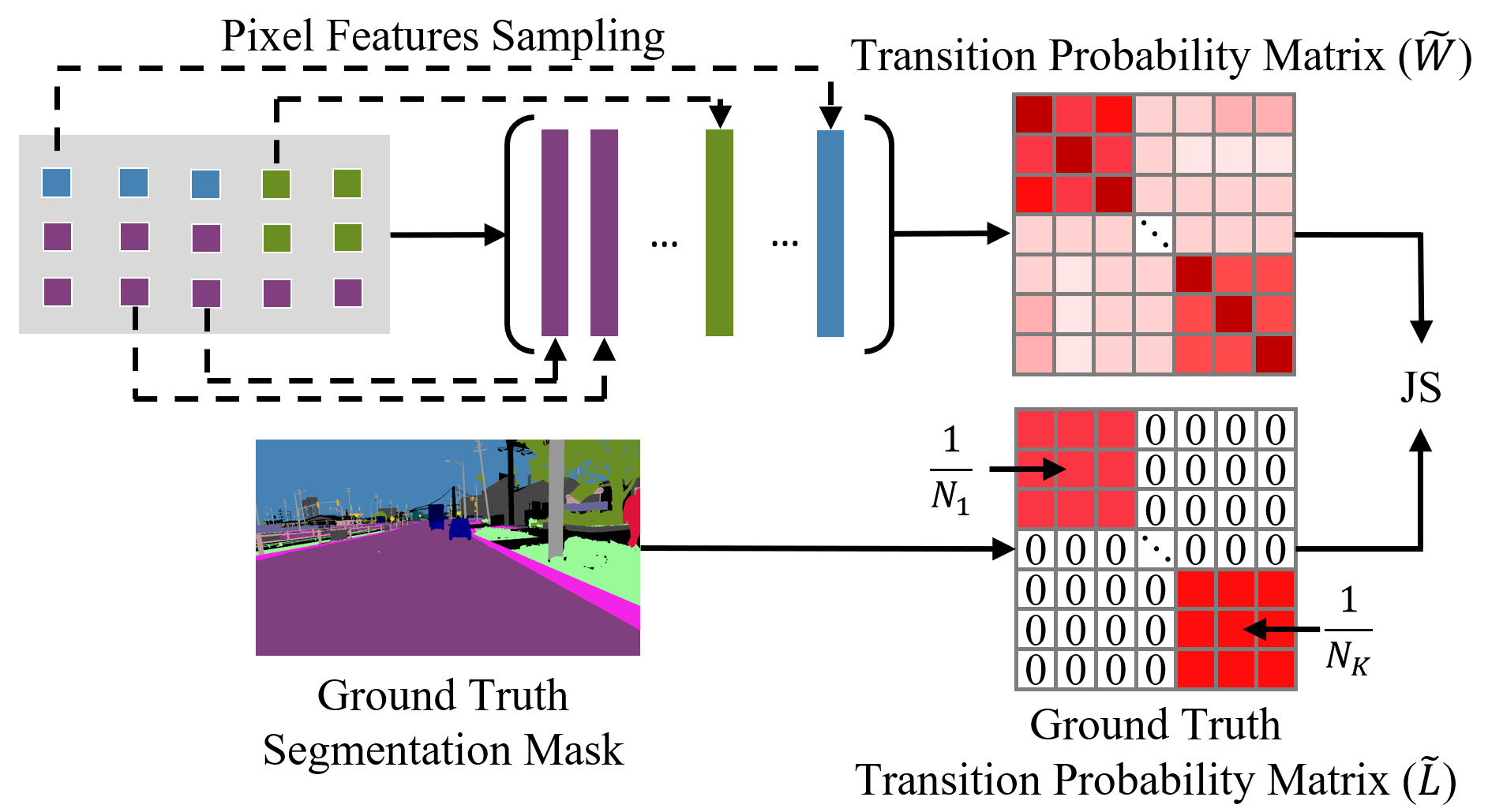}
	\caption{Illustration of pixel-to-pixel contrastive learning.}
	\label{contrastive_pic}
\end{figure}

\textbf{Pixel-to-Pixel Contrastive Learning.}
This learning loss is to constrain the pixels in feature maps having the same class label should be closer and different class labels should be distant in the feature space. To realize this goal, we propose a novel pixel-to-pixel contrastive loss. As shown in Fig.~\ref{contrastive_pic}, for the $k$-th class, we first sample $N_k$ features to avoid memory explosion when using all pixels in the feature maps. We denote $N$ as the number of features sampled in a batch, \ie $N=\sum_{k=1}^KN_k$, $K$ is class number. We next calculate the similarity matrix $W\in\mathbb{R}^{N\times N}$ over the sampled pixel-wise features, in which $W_{ij}={\mathrm{exp}}(z_i\cdot z_j/\tau)$, ``$\cdot$'' denotes inner product, $\tau$ is temperature. We can also get the ground truth label matrix $L$, which implies the semantic relationship among sampled pixels, \ie $L_{ij}=1$ if $y_{z_i}=y_{z_j}$ else $L_{ij}=0$. Then, we can calculate the transition probability matrix $\widetilde{W}$ and $\widetilde{L}$ by normalizing each row of $W$ and $L$:
\begin{equation}\label{normalize_row_inpixeltopixel}
    \widetilde{W}=D_W^{-1}W, \widetilde{L}=D_L^{-1}L
\end{equation}
where $D_W={\mathrm{diag}}(W\Vec{1}),D_L={\mathrm {diag}}(L\Vec{1})$. We define pixel-to-pixel contrastive loss by calculating distribution distance in each row between $\widetilde{W}$ and $\widetilde{L}$:
\begin{equation}\label{loss2}
    \mathcal{L}_{pp} = \frac{1}{N}\sum_{i=1}^N {\mathcal{M}}(\tilde{w}_i,\tilde{l}_i)
\end{equation}
where $\tilde{w}_i$ and $\tilde{l}_i$ are the $i$-th row in matrix $\widetilde{W}$ and $\widetilde{L}$. Both $\tilde{w}_i$ and $\tilde{l}_i$ in Eq.~\eqref{loss2} are in probability simplex, and $\mathcal{M}(\cdot)$ is the distribution distance metric. In this paper, we adopt JS divergence as metric $\mathcal{M}(\cdot)$, which is a symmetric divergence. Note that our pixel-to-pixel semantic similarity loss is different from the supervised contrastive learning loss \cite{scl} in two aspects. Firstly, our loss considers the feature with itself as positive pair, positioning along the diagonal of the row normalized matrix $\widetilde{W}$. Secondly, we use a symmetric JS divergence as our distribution metric. In the experiment, we will show that our proposed pixel-to-pixel contrastive loss produces better generalization performance than the standard supervised contrastive loss. 

\textbf{Pixel-to-Class Contrastive Learning.} This loss is to enforce that pixel-level features in the feature maps should be closer to their own class centers, represented by class prototypes. We introduce our pixel-to-class similarity loss as 
\begin{equation}\label{loss3}
\mathcal{L}_{pc}=\frac{1}{H^{\prime}W^{\prime}}\sum_{i=1}^{H^{\prime}W^{\prime}}-y^T_{z_i}{\mathrm {log}}\frac{{\mathrm {exp}}(z_i\cdot P^{k}/\tau)}{\sum_{P^a \in \mathcal{P}}{\mathrm {exp}}(z_i\cdot P^a/\tau)}
\end{equation}
where $\mathcal{P}$ is the set of class prototypes. Specifically, we use class prototype $P$ before updating in the current batch to calculate pixel-to-class contrastive loss. In fact, pixel-to-class contrastive loss is a standard classification loss, to ensure each pixel can be classified by the class prototype classifier. 

\textbf{Instance-to-Class Contrastive Learning.} In addition to the above losses, we also constrain that the class prototype can correctly classify each instance prototype, which is computed by average pooling features in each connected region of each class. We can use contrastive loss like Eq.~\eqref{loss3}, however, toughly pulling all different instance prototypes of a class closer to the class prototype may lose the diversity of instance-level feature of the class. We adopt the margin triplet loss \cite{triplet} as our instance-to-class contrastive learning loss:
\begin{footnotesize}
\begin{equation}\label{loss4}
{\mathcal{L}_{ic}=\frac{1}{K}\sum_{k=1}^K\frac{1}{M_k}\sum_{m,n}\max\{d(p_m^k,P^k)+\xi-d(p_n^{\setminus k},P^k),0\}}
\end{equation}
\end{footnotesize}where $p_m^k$ is $m$-th instance prototype in $k$-th class, $p_n^{\setminus k}$ is the $n$-th instance prototype in all classes except $k$, $M_k$ is the total number of instance pairs for class $k$, $\xi$ is the margin, $d$ is $L_2$ distance. We can use Eq.~\eqref{equ4} to get each instance prototype by substituting the class mask with instance binary mask. Each binary mask is accessed by extracting the connected region in each class mask as \cite{da5}.

\textbf{Multi-level Contrastive Loss.} Totally, our multi-level contrastive loss is defined as
\begin{equation}\label{loss6}
\mathcal{L}_{mlcl}=\lambda\mathcal{L}_{pp}+\mathcal{L}_{pc}+\mathcal{L}_{ic}
\end{equation}
where we only have one hyper-parameter $\lambda$ in the loss to balance the contribution of pixel-to-pixel contrastive loss.

\subsection{Training methods}
The training phase of our approach consists of two stages. First, we use Eq.~\eqref{loss1} to pre-train our SSDP network by reconstructing the original data from augmented data. In the second stage, we freeze the parameters of SSDP and only use it for data projection. Based on the projected data, except for task loss and multi-level contrastive loss, we utilize a divergence loss to make class prototypes apart from each other after each update, which is denoted as
\begin{equation}\label{loss5}
    \mathcal{L}_{div}= \frac{1}{K(K-1)}\sum_{j=1}^K\sum_{i\neq j}^K max\{\hat{P}^i(\hat{P}^j)^T,0\}
\end{equation}
Finally, we use the following total loss to train our segmentation model based on projected data in the second stage
\begin{equation}\label{loss6}
\mathcal{L}_{total}=\mathcal{L}_{task}+ \mathcal{L}_{mlcl}+\mathcal{L}_{div}
\end{equation}
we use a per-pixel cross-entropy loss for segmentation task loss  $\mathcal{L}_{task}$. To avoid feature mode collapse by using $\mathcal{L}_{mlcl}$ at the beginning, we warm up our segmentation model only using $\mathcal{L}_{task}$ for ten epochs and then use $\mathcal{L}_{total}$ to train. 
\begin{table*}[t]
\centering
\begin{tabular}[t]{c|c|cccc|cccc}
    \hline 
    \multirow{2}{*}{Backbone}& \multirow{2}{*}{Method} &\multicolumn{4}{c|}{Train on GTAV (G)}& \multicolumn{4}{c}{Train on Cityscapes (C)}\\ 
    \cline{3-10}
     ~&~ & C & B & M & Mean & B & S & G & Mean\\
     \hline
\multirow{12}*{ResNet50}
~ & Baseline &28.95 &25.12 & 28.18 & 27.42 
&44.91 &23.29 & 42.55 &36.92 \\
~ & SW & 29.91 & 27.48 & 29.71 & 29.03 
&48.49 &26.10 &44.87 &39.82\\
~ & IBN-Net &33.85 &32.30 &37.75 &34.63
&48.56 &26.14 &45.06 &39.92 \\
~ & DPRC &37.42 &32.14 &34.12 &34.56
&49.86&26.58& 45.62 & 40.69\\
~ & GTR &37.53 &33.75 &34.52 &35.27
&50.75 & 26.47&45.79 & 41.00\\
~ & IRW &33.57 &33.18 &38.42 &35.06
& 48.67 & 26.05 & 45.64 & 40.12\\
~ & ISW &36.58 &35.20 &40.33 &37.37
&50.73 & 26.20& 45.00 & 40.64\\
~ & SANSAW & 39.75 & 37.34 & \underline{41.86} &39.65 
&\textbf{52.95} & \textbf{28.32} & \textbf{47.28} &\textbf{42.85} \\
~ & DIRL & \underline{41.04} &\underline{39.15}&41.60&\underline{40.60}
&51.80 & 26.50 & \underline{46.52} & 41.60 \\
~ & \textbf{DPCL} & \textbf{44.87}&\textbf{40.21} &\textbf{46.74}  & \textbf{43.94} 
& \underline{52.29} & \underline{26.60} & 46.00 & \underline{41.63} \\
\cline{2-10}
~ & \textbf{DPCL+TTA (C)}& \textbf{46.34} &40.67 &48.28& 45.10
&52.23 &26.68&46.26 &41.72\\

~ & \textbf{DPCL+TTA (C+E)}&
46.02 &\textbf{41.14} &\textbf{48.79} & \textbf{45.32}
&\textbf{53.30} &26.91 &47.25 & 42.49\\

\hline

\multirow{7}*{ShuffleNetV2}
~ & Baseline &25.56 &22.17 & 28.60 & 25.44
&38.09 &21.25 & 36.45 &31.93\\
~ & IBN-Net &27.10 &31.82 &34.89 &31.27
&41.89 &22.99 &40.91& 35.26\\
~ & ISW &30.98 &32.06 &35.31 &32.78
& 41.94 & 22.82& 40.17 & 34.98\\
~ & DIRL & \underline{31.88} &\underline{32.57}& \underline{36.12} & \underline{33.52}
& \underline{42.55} & \textbf{23.74} & \underline{41.23} &\underline{35.84}\\
~ & \textbf{DPCL} &  \textbf{36.66}& \textbf{34.35} & \textbf{39.92} & \textbf{36.98}
& \textbf{43.96} & \underline{23.24}&\textbf{41.93} &\textbf{36.38}\\
\cline{2-10}
~ & \textbf{DPCL+TTA (C)}&\textbf{39.12} &\textbf{35.86} &\textbf{42.19} & \textbf{39.06}
&44.18 &23.60 &42.23 & 36.67\\
~ & \textbf{DPCL+TTA (C+E)}&
37.94 &35.40 &41.15 &38.16
&\textbf{44.53} &\textbf{23.95} &\textbf{43.49} & \textbf{37.32}\\
\hline
\multirow{7}*{MobileNetV2}
~ & Baseline &25.92 &25.73 & 26.45 & 26.03
&40.13&21.64&37.32&33.03\\
~ & IBN-Net &30.14 &27.66 &27.07 &28.29
&44.97&23.23&41.13&36.44\\
~ & ISW &30.86 &30.05 &30.67 &30.53
&45.17&22.91&41.17&36.42\\
~ & DIRL & \underline{34.67}&\underline{32.78}&\underline{34.31}&\underline{33.92}
&\textbf{47.55}&\underline{23.29}&\underline{41.43}&\underline{37.42}\\
~ & \textbf{DPCL} & \textbf{37.57} & \textbf{35.45} & \textbf{40.30}&\textbf{37.77}
&\underline{46.23}&\textbf{24.68}&\textbf{44.17}&\textbf{38.36}\\
\cline{2-10}
~ & \textbf{DPCL+TTA (C)}&\textbf{41.16} &36.59 &\textbf{42.94} & \textbf{40.23} &46.37 &24.76&44.32 & 38.48\\
~ & \textbf{DPCL+TTA (C+E)}&
39.13 &\textbf{36.86} &41.83 & 39.27
&46.76 &\textbf{25.17} &\textbf{45.49} & \textbf{39.14}\\
\hline
\end{tabular}
\caption{Results for the task G to C, B and M and the task C to B, S and G in mIoU. The best and second best results of methods without TTA are bolded and underlined respectively. The best TTA methods are also bolded.} \label{compare_task}
\end{table*}

\begin{figure*}[htbp]   
	\centering
	\includegraphics[width=
 0.9\linewidth,scale=1.00]{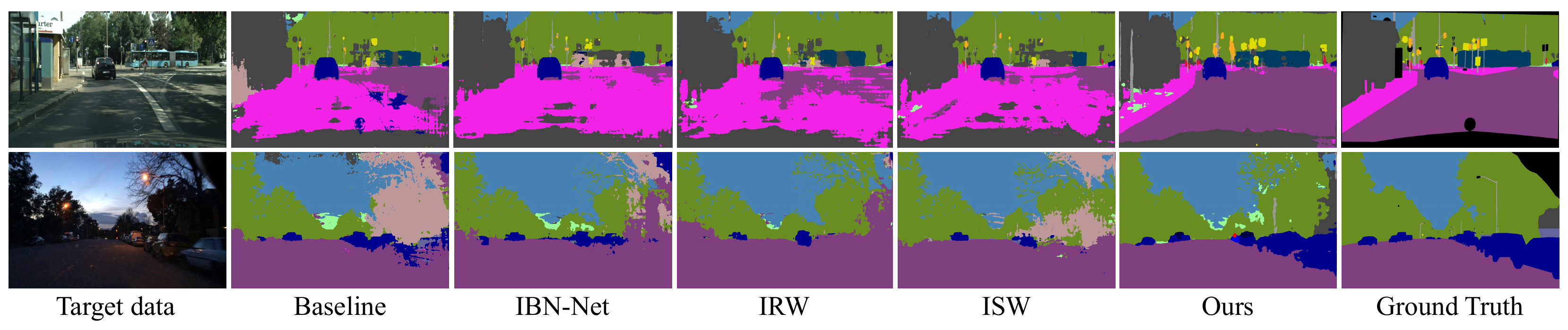}
	\caption{Visual comparison of different domain generalization semantic segmentation methods using ResNet50, trained on GTAV (G) and tested on unseen target domains of Cityscapes (C)~\cite{cityscapes} and BDD (B)~\cite{bdd100k}.}
	\label{Qualitative}
\end{figure*}

\subsection{Testing process}
In testing, we first project target data by our SSDP to mitigate domain shift. Then we send the projected data into segmentation model to generate its prediction for segmentation. Our class prototypes obtained in the training phase can also be regarded as a classifier. So we average the softmax probabilities predicted by classifier $H$ and class prototypes (based on features in the second last layer) to make a more reliable prediction. Except for standard test process, we also propose a test-time adaptation scheme by minimizing our proposed pixel-to-pixel contrastive loss. Different from existing test-time adaptation methods \cite{tent,adacontrast}, which commonly update model parameters in the test time process, we fix all the network parameters, and only optimize the input image of segmentation network, taking the projected data by SSDP as initialization. Specifically, given a projected target domain data by SSDP, we first compute its pseudo label by averaging the predictions from classifier $H$ and class prototypes, then we randomly sample one thousand pixel of each class from this image without replacement to construct our pixel-to-pixel contrastive loss in Eq.~\eqref{loss2}, we iterate the projected data once by gradient descent to minimize the loss, and get the refined prediction of segmentation mask.

\section{Experiment}
In this section, we will evaluate our method on different domain generalization benchmarks.
\subsection{Experimental Setups}
\textbf{Synthetic Datasets}. GTAV~(G) \cite{gtav} is a synthetic dataset, which contains 24966 images with resolution of $1914\times1052$ along with their pixel-wise semantic labels, and it has 12,403, 6,382, and 6,181 images for training, validation, and test, respectively. SYNTHIA~(S) \cite{synthia} is an another synthetic dataset. The subset SYNTHIA-RANDCITYSCAPES is used in our experiments which contains 9400 images with resolution of $1280\times760$.

\noindent
\textbf{Real-World Datasets.} Cityscapes~(C) \cite{cityscapes} is a high resolution dataset ($2048\times1024$) of 5000 vehicle-captured urban street images taken from 50 cities primarily in Germany. BDD~(B) \cite{bdd100k} is another real-world dataset that contains diverse urban driving scene images in resolution of $1280\times720$. The last real-world dataset we use is Mapillary~(M) \cite{mapillary}, which consists of 25,000 high-resolution images with a minimum resolution of $1920\times1080$ collected from all around the world.

\noindent
\textbf{Implementation Details.}
We use ResNet50 \cite{ResNet}, ShuffleNetV2 \cite{shufflenetv2} and MobileNetV2 \cite{mobilenetv2} as our segmentation backbones for the task GTAV to Cityscapes, BDD and Mapillary and the task Cityscapes to BDD, SYNTHIA and GTAV.  We take SGD optimizer with an initial learning rate of 1e-3, and train segmentation model for 40k iterations with batch size of 8, momentum of 0.9 and weight decay of 5e-4. We adopt the polynomial learning rate scheduling \cite{parsenet} with the power of 0.9. We use color-jittering and Gaussian noise as image augmentation. We also use random cropping, random horizontal flipping, and random scaling to avoid the model over-fitting. As for our SSDP subnet, we adopt the same architecture with generator in CycleGAN \cite{cyclegan} and train it with Adam optimizer. We utilize the same image data augmentation with our segmentation network. In the multi-level contrastive learning, we sample thirty pixel features in each class from a batch of images, half of which are with incorrect prediction by segmentation classifier and half of which are with correct prediction according to their labels. We respectively use $q=10$ and $q=5$ for the task trained on GTAV and Cityscapes. The other parameters are set as $\xi=0.5, \tau=0.1, \lambda=5$.

\noindent
\textbf{Compared methods.} Our baseline model is DeepLabV3+ trained by cross-entropy loss in source domain for segmentation. We compare with the DG methods: SW \cite{sw}, IBN-Net \cite{ibn}, DPRC \cite{DPRC}, GTR \cite{gtr}, IRW, ISW \cite{isw}, DIRL \cite{dirl} and SANSAW \cite{snsw}.

\subsection{Comparison with state-of-the-art methods}
As for the synthetic to real generalization, we follow DIRL \cite{dirl} to evaluate the generalization performance from GTAV to Cityscapes, BDD and Mapillary. As shown in Table~\ref{compare_task}, our method outperforms the other methods clearly and consistently across three different network backbones, especially for the task from GTAV to Cityscapes and Mapillary. When using ResNet50, our method improves performance from 41.04 to 44.87 on Cityscapes dataset and from 41.60 to 46.74 on Mapillary dataset compared with DIRL. 
Except for standard test process, we also show our method performance with test-time adaptation as discussed in the subsection of Testing process, which is denoted as DPCL+TTA (C). 
As for the backbone of ShuffleNetV2 and MobileNetV2, our method respectively improves the mIoU by 2.08 and 2.46 using test-time adaptation. Except for contrastive loss, we additionally try TTA  by minimizing sum of entropy and our pixel-to-pixel contrastive loss (normalized by number of selected pixels), dubbed DPCL+TTA (C+E), and it further improves performance with ResNet50. We also visualize the qualitative comparisons with other methods shown in  Fig.~\ref{Qualitative} to show superiority of our methods DPCL. 

\begin{table}[t!]
\centering
{\begin{tabular}[t!]{ccc|ccc|c}
\hline
 SSDP & $\mathcal{L}_{mlcl}$& $\mathcal{L}_{div}$& C & B & M & Mean\\
\hline
 && &  28.95 &25.12 & 28.18 & 27.42\\
 \checkmark && &40.13 &39.47 & 43.13 & 40.91 \\
 \checkmark&\checkmark& & 43.68& 39.89 & 45.05&42.87 \\
 \checkmark &\checkmark&\checkmark& 44.87&40.21 &46.74  & 43.94 \\
\hline
\end{tabular}}
\caption{Ablation study for domain generalization task G to C, B and M with ResNet50 in mIoU, SSDP denotes our Self-supervised Source Domain Projection network, $\mathcal{L}_{mlcl}$  denotes Multi-level Contrastive Learning loss and $\mathcal{L}_{div}$ denotes class prototype divergence loss.}\label{ablation}
\end{table}


We further compare our methods with other methods from Cityscapes to BDD, SYNTHIA and GTAV, shown in Table~\ref{compare_task}. Our method achieves the best performance with backbone ShuffleNetV2 and MobileNetV2, achieves the second best performance with backbone ResNet50 among the compared methods. Our method DPCL+TTA (C+E) further improves the performance for three backbones. 


\subsection{Ablation Study}
We examine each component of our method DPCL to check how they contribute in the domain generalization on the task GTAV to Cityscapes, BDD and Mapillary. As show in Table~\ref{ablation}, the baseline method shows lowest performance on three unseen target domains. Our method improves baseline  in average accuracy from 27.42 to 40.91 by using our SSDP. This shows that our SSDP for projecting data can mitigate domain shift in the test phase. Based on the projected data, we add our multi-level contrastive learning module, further improving the performance. Finally, we add diversity constraint $\mathcal{L}_{div}$ to our class prototypes and produce the best performance, especially in the task GTAV to Mapillary. 

\begin{table}[t!]
\centering
\begin{tabular}[t!]{c|ccc|c}
\hline
Method & C & B & M & Mean\\
\hline
SSDP~(w/~AE) &36.41 &34.47 & 36.61 & 35.83\\
SSDP~(w/o~AdaIN) &43.13 &39.77 &45.10 & 42.67 \\
SSDP &44.87 &40.21 & 46.74 & 43.94 \\
\hline
\end{tabular}
\caption{Results of different designs of Source Domain Projection network for task G to C, B and M using ResNet50.}\label{ae_compare}
\end{table}

\begin{table}[t]
\centering
\begin{tabular}[t]{c|cccc}
\hline
Method & Scl-CE & Scl-JS &Ours-CE & Ours-JS\\
\hline
Mean mIoU& 43.08 & 43.65 &  43.46 & 43.94 \\
\hline
\end{tabular}
\caption{Results of different choices of pixel-to-pixel contrastive loss for task G to C, B and M using ResNet50. Mean mIoU is obtained over the three target dataset.
}\label{pixeltopixelloss}
\end{table}

\noindent
\textbf{Comparison of different designs of SSDP.}
We compare different designs of SSDP shown in Table~\ref{ae_compare}. In the first row, we use a standard Auto-Encoder in SSDP network, denoted as SSDP~(w/ AE), which aims to reconstruct original input image and obtains 35.83 mean mIoU. In the second row, we input augmented data into SSDP and directly reconstruct original data without AdaIN technique in the feature space named SSDP~(w/o AdaIN). It improves the average performance from 35.83 to 42.67, which is superior than DIRL. The last row is the SSDP that we adopt, which reconstructs the original data from augmented data with AdaIN technique \cite{adain} in the feature space and shows effectiveness in average performance.

\noindent
\textbf{Comparison of different choices of pixel-to-pixel contrastive learning.} In this paragraph, we compare our pixel-to-pixel contrastive loss with supervised contrastive loss \cite{scl} under the same hyper-parameter setting. The method Scl-CE is the standard supervised contrastive loss used in \cite{scl}. Compared with ours, Scl-CE discards the diagonal values of $W$ and $L$ and uses cross-entropy loss (see appendix). Scl-JS masks out the diagonal vector of matrix $W$ and $L$, but uses JS divergence as metric $\mathcal{M}(\cdot)$.  Ours-CE and Ours-JS are respectively our loss using cross-entropy and JS divergence as metric $\mathcal{M}(\cdot)$. Table~\ref{pixeltopixelloss} shows that Ours-JS achieves consistently better performance than the other variants of losses.

Due to space limit, more visualization results and empirical analysis, \eg sensitivity to hyper-parameters, ablation for multi-level contrastive learning loss, data augmentation, choices of $\mathcal{L}_{ic}$, etc., are in the appendix .

\section{Conclusion}

In this paper, we propose a novel domain generalization semantic segmentation method DPCL, consisting of modules of Self-supervised Source Domain Projection (SSDP) and Multi-level Contrastive Learning (MLCL). 
Comprehensive experiments demonstrate the effectiveness of SSDP and MLCL in domain generalization semantic segmentation. In the future, we plan to further improve the learning schemes on the segmentation model, and try transformer-based  backbones in our framework.
\appendix

\renewcommand{\thetable}{A-\arabic{table}}
\renewcommand{\theequation}{A-\arabic{equation}}
\renewcommand{\thefigure}{A-\arabic{figure}}
\setcounter{table}{0}
\setcounter{figure}{0}
\setcounter{equation}{0}
\section{A. Comparison of our pixel-to-pixel contrastive loss with supervised contrastive loss}
In this section, we will compare our transition probability matrix-based pixel-to-pixel contrastive loss with the supervised contrastive learning loss used in \cite{scl}. 

\textbf{Supervised contrastive loss}. The supervised contrastive learning \cite{scl} is defined as
\begin{equation}\label{sclloss}
    \mathcal{L}_{sup}=\sum_{i\in N}\frac{-1}{|\mathcal{P}^+(i)|}\sum_{p\in \mathcal{P}^+(i)}{\mathrm{log}}\frac{\mathrm{exp}(z_i\cdot z_p/\tau)}{\sum\limits_{j\in \mathcal{A}(i)}\mathrm{exp}(z_i\cdot z_j/\tau)}
\end{equation}
where $N$ is index set of samples, $\mathcal{A}(i)=N\setminus\{i\}$, $\mathcal{P}^+(i) = \{p\in \mathcal{A}(i): y_{z_p}=y_{z_i}\}$ is the set of indices of all positive samples of sample $i$, $y_{z_p}$ and $y_{z_i}$ are the class labels of sample $z_p$ and $z_i$, $|\mathcal{P}^+(i)|$ is the cardinality of set $\mathcal{P}^+(i)$, ``$\cdot$'' is inner product, $\tau$ is temperature parameter. 

\textbf{Our transition probability matrix-based contrastive loss}. We define our pixel-to-pixel contrastive loss based on the transition probability matrix. We first calculate the similarity matrix $W$ among samples, in which $W_{ij}={\mathrm{exp}}(z_i\cdot z_j/\tau)$ and get the ground truth label matrix $L$, which implies the semantic relationship among samples, \ie $L_{ij}=1$ if $y_{z_i}=y_{z_j}$ else $L_{ij}=0$. Then we separately calculate transition probability matrix  $\widetilde{W}$ and $\widetilde{L}$ by normalizing each row of $W$ and $L$ using Eq.~\eqref{normalize_row_inpixeltopixel}, and finally get our transition probability matrix-based pixel-to-pixel contrastive loss as
\begin{equation}\label{oursjs}
    \mathcal{L}_{pp} = \frac{1}{N}\sum_{i=1}^N {\mathcal{M}}(\tilde{w}_i,\tilde{l}_i)
\end{equation}
where $\tilde{w}_i$ and $\tilde{l}_i$ are $i$-th rows in matrix $\widetilde{W}$ and $\widetilde{L}$. $\mathcal{M}(\cdot)$ is the distribution distance metric. In this paper, we adopt JS divergence as metric $\mathcal{M}(\cdot)$, which is a symmetric divergence

\textbf{Comparison of two losses.} Next we will illustrate the differences between supervised contrastive loss \cite{scl} and our transition probability matrix-based contrastive loss in two aspects. Firstly, for supervised contrastive loss in Eq.~\eqref{sclloss}, the positive sample set of sample $i$ does not include sample $i$ itself. In our transition probability matrix-based contrastive loss, we consider the sample $i$ itself as the positive sample of $i$, as in the diagonal vector of $W$. Secondly, we use JS divergence as our metric $\mathcal{M}(\cdot)$, which is a symmetric divergence to calculate the distance between our $\tilde{w}_i$ and $\tilde{l}_i$, which is different form standard cross entropy used in \cite{scl}. In the experiment, 
the mIoU 
for using supervised contrastive loss is 43.08. The mIoU is improved to 43.46 by including the sample itself as the positive paired sample, as in Table~\ref{pixeltopixelloss}. By further using the JS divergence to replace the cross-entropy loss which constructs our pixel-to-pixel contrastive loss, the mIoU increases to 43.94, as shown in Table~\ref{pixeltopixelloss}.

\begin{figure}[t!]   
	\centering
	\includegraphics[width=\linewidth,scale=1.00]{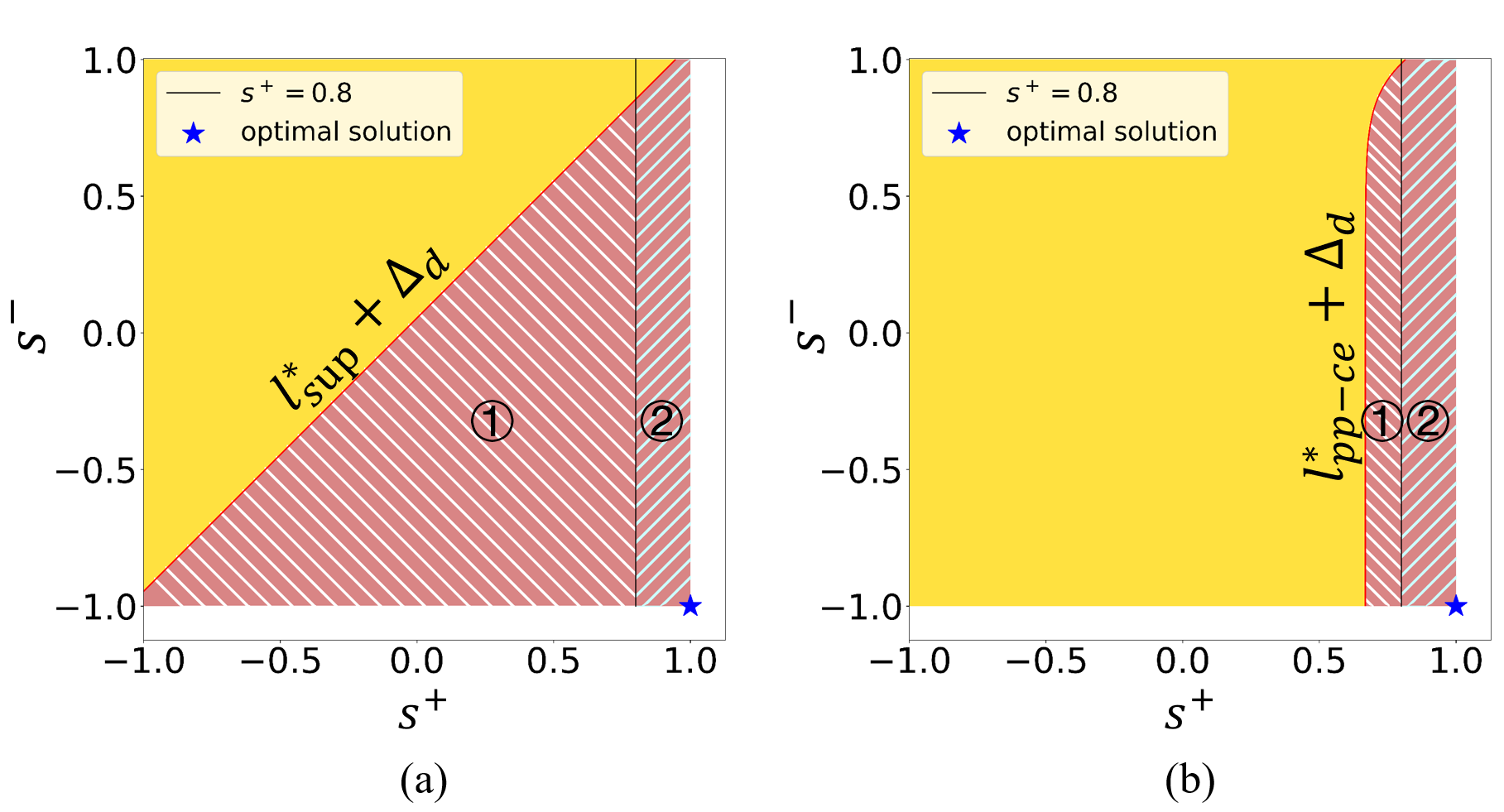}
	\caption{Comparison of our transition probability matrix-based pixel-to-pixel contrastive loss and supervised contrastive loss. (a) The loss contour with value of $l^*_{sup}+\Delta_d$ for the loss $\mathcal{L}_{sup}$. The $l^*_{sup}$ is the minimal value of loss $\mathcal{L}_{sup}$. (b) The loss contour with value of $l^*_{pp-ce}+\Delta_d$ for the loss $\mathcal{L}_{pp-ce}$. The $l^*_{pp-ce}$ is the minimal value of loss $\mathcal{L}_{pp-ce}$. We adopt $\tau=0.1, \Delta_d=1$ for both (a) and (b).}
	\label{compare_loss}
\end{figure}

To further illustrate the benefit of using the sample itself as the paired positive sample. We next give an example with three samples to illustrate the differences between $\mathcal{L}_{sup}$ and our pixel-to-pixel contrastive loss using cross-entropy as the metric $\mathcal{M}(\cdot)$, which is denoted as $\mathcal{L}_{pp-ce}$. Specifically, 
we denote the anchor sample as $z_a$, its positive sample as $z_p$ and negative sample as $z_n$. All the samples are $l_2$ normalized. We denote $s^+=z_a\cdot z_p, s^-=z_a\cdot z_n, s_a = z_a\cdot z_a=1$. Then, the supervised contrastive loss \cite{scl} can be written as
\begin{equation}
    \mathcal{L}_{sup}=-\mathrm{log}\frac{\mathrm{exp}(s^+/\tau)}{\mathrm{exp}(s^+/\tau)+\mathrm{exp}(s^-/\tau)}.
\end{equation}
Meanwhile, our pixel-to-pixel contrastive loss with cross-entropy is written as
\begin{equation}
    \begin{split}
        \mathcal{L}_{pp-ce}=-&\frac{1}{2}
        \left( \mathrm{log}\frac{\mathrm{exp}(s_a/\tau)}{\mathrm{exp}(s_a/\tau)+\mathrm{exp}(s^+/\tau)+\mathrm{exp}(s^-/\tau)}\right.\\
        &\left.+\mathrm{log}\frac{\mathrm{exp}(s^+/\tau)}{\mathrm{exp}(s_a/\tau)+\mathrm{exp}(s^+/\tau)+\mathrm{exp}(s^-/\tau)} 
        \right)
    \end{split}
\end{equation}
Note that the minimization of losses $\mathcal{L}_{sup}$ and $\mathcal{L}_{pp-ce}$ are both achieved when sample similarity $s^+$ is 1 and negative sample similarity $s^-$ is -1. Though their minimizers (\ie optimal solutions) are the same, we next empirically show that the approximated optimal solutions, \ie the solutions with loss values within an error bound to the minimal loss value, obtained by our proposed $\mathcal{L}_{pp-ce}$ could be better. To do this, we respectively visualize the loss contours of level $l^*+\Delta_d$ shown by the red line in Fig.~\ref{compare_loss}, where $l^*$ is the minimum loss value, $\Delta_d$ is the distance away from the minimum loss value. As shown in Fig.~\ref{compare_loss}, the area below the red line denotes the set of approximate optimal solutions within $\Delta_d$ error bound to the minimal loss, and the optimal solution is the blue start point in the right bottom corner. The shaded areas \ding{172} and \ding{173}  are the regions in which the approximated optimal solutions with positive sample similarity $s^+$ smaller and larger than 0.8, respectively. 
The ratio of area \ding{173} to the sum of area \ding{172} and \ding{173} means the conditional probability of positive sample similarity larger than 0.8 given that the loss is smaller than $l^*+\Delta_d$. In the Fig.~\ref{compare_loss}, we can see that the ratio of area \ding{173} to the sum of area \ding{172} and \ding{173} obtained by $\mathcal{L}_{pp-ce}$ is apparently larger than that obtained by $\mathcal{L}_{sup}$. This implies that the approximated solution of $\mathcal{L}_{pp-ce}$ is more possible that the positive samples are highly similar, which is expected in contrastive learning. This analysis may account for the better results obtained by our pixel-to-pixel contrastive loss.

\begin{table}[t]
\centering
\begin{tabular}[t]{c|c}
\hline
Method &mIoU\\
\hline
No adaptation &44.87  \\
Tent &45.80 \\
\textbf{DPCL+TTA (E)} & 45.90\\
\textbf{DPCL+TTA (C)}  & \textbf{46.34}\\
\hline
\end{tabular}
\caption{Results of different test-time adaptation methods for task GTAV \cite{gtav} to Cityscapes \cite{cityscapes} using ResNet50.  The best result is bolded.}\label{testtimetask}
\end{table}
\begin{table}[t]
\centering
\begin{footnotesize}

\begin{tabular}[t]{c|ccccc}
\hline
Iteration &1&2 &3& 4&5 \\
\hline
Tent &45.80 & 45.99 & 45.88 &  45.61&45.28 \\
DPCL+TTA (C) &46.34 & 46.40 & 46.20 &45.96 &45.70\\
\hline
\end{tabular}
\end{footnotesize}

\caption{Results of different iteration times of Tent and DPCL+TTA (C) for task GTAV \cite{gtav} to Cityscapes \cite{cityscapes} using ResNet50.}\label{differentiterationnums}
\end{table}

\section{B. Comparison of different test-time adaptation strategies}
In this section, we compare different test-time adaptation methods for the task GTAV \cite{gtav} to Cityscapes \cite{cityscapes} with ResNet50. Given an unlabel target data, we first project it by our Self-supervised Source Domain Projection (SSDP) and next we use different test-time adaptation strategies  based on the projected data. As shown in Table~\ref{testtimetask}, the first row is the result of our method without test-time adaptation. Tent \cite{tent}, which updates parameters of BN layers by minimizing entropy loss, improves the results from 44.87 to 45.80. The third and forth rows are our test-time adaptation strategies that iterate the projected data rather than updates the trained model's parameters. The third row is the result of our test-time adaptation approach DPCL+TTA (E) that updates the input data by minimizing entropy loss with gradient descent. The last row is the result of our approach DPCL+TTA (C) that update the input data by minimizing our transition probability matrix-based pixel-to-pixel contrastive loss. Our method DPCL+TTA (C) achieves the best performance among compared methods.

We further compare the results for different iterations of Tent \cite{tent} and our method DPCL+TTA (C) as shown in Table~\ref{differentiterationnums}. Both our method and Tent \cite{tent} get the best performance when iterating twice. In order to save time, we only iterate once in our test-time adaptation experiment.

\section{C. More quantitive results.}
In this section. we show more quantitive results of our method using ResNet50. Unless otherwise specified, our results are the average mIoU of the task GTAV (G) \cite{gtav} to Cityscapes (C) \cite{cityscapes}, BDD (B) \cite{bdd100k} and Mapillary (M) \cite{mapillary}.

\noindent
\textbf{Ablation study.} 
We investigate each component in Multi-level Contrastive Learning (MLCL) module to research how they contribute to our total method. As shown in Table~\ref{ablation_mlcl}, each of contrastive loss is effective for the total performance. Our pixel-to-pixel contrastive loss based on Transition Probability Matrix contributes the most.

\begin{table}[t]
\centering
{\begin{tabular}[t]{c|c}
\hline
Method &Mean mIoU\\
\hline
 DPCL (w/o $\mathcal{L}_{pp}$)&41.14\\
 DPCL (w/o $\mathcal{L}_{pc}$&42.12\\
 DPCL (w/o $\mathcal{L}_{ic}$)&43.12\\
 DPCL&43.94\\
\hline
\end{tabular}}
\caption{Ablation study for each part in Multi-level Contrastive learning loss for domain generalization task G to C, B and M with ResNet50, Mean mIoU means the  average mIoU over three subtasks. $\mathcal{L}_{pp}$ denotes our pixel-to-pixel contrastive loss, $\mathcal{L}_{pc}$ denotes our pixel-to-class contrastive loss and $\mathcal{L}_{ic}$ denotes our  instance-to-class contrastive loss.}\label{ablation_mlcl}
\end{table}

\noindent
\textbf{Hyper-parameter Impacts.} 
We investigate the effect of parameter $\lambda$ in our multi-level contrastive loss, marginal parameter $\xi$, temperature parameter $\tau$ and number of cluster centers $q$ on our method. 
As shown in Table~\ref{lamda_choice},  our method is relatively stable when $\lambda$ changing from 3 to 7. And the performances are consistently better than DIRL \cite{dirl}. Moreover, varying $\xi$ from 0.1 to 0.9 does not have a significant impact on our method's performance, and we observe the best results when $\xi$ is set to 0.7 as shown in Table~\ref{ablation_xi}. We also investigate the effect of the temperature parameter $\tau$, and find that our method performs best when $\tau$ is set to 0.1, as presented in Table~\ref{ablation_tau}. Lastly, we examine the impact of the number of cluster centers $q$ varying from 5 to 30 and find that our method exhibits relative stability in performance as q increased, as shown in Table~\ref{ablation_q}.

\begin{table}[t!]
\centering
\begin{tabular}[t!]{c|ccccc}
\hline
$\lambda$ & 3 &4& 5&6& 7\\
\hline
Mean mIoU& 43.22&43.53& 43.94&43.07 &43.59 \\
\hline
\end{tabular}
\caption{Sensitivity to parameter $\lambda$ on task G to C, B and M using ResNet50.}\label{lamda_choice}
\end{table}

\begin{table}[t!]
\centering
\begin{tabular}[t!]{c|ccccc}
\hline
$\xi$ & 0.1 &0.3& 0.5&0.7& 0.9\\
\hline
Mean mIoU& 43.60&43.50&43.94&43.96&43.77 \\
\hline
\end{tabular}
\caption{Sensitivity to parameter $\xi$ on task G to C, B and M using ResNet50.}\label{ablation_xi}
\end{table}

\begin{table}[t]
\centering
\begin{tabular}[t]{c|ccccc}
\hline
$\tau$ & 0.05 &0.1& 0.2&0.3& 0.5\\
\hline
Mean mIoU& 43.02&43.94&42.47&41.74&40.22 \\
\hline
\end{tabular}
\caption{Sensitivity to parameter $\tau$ on task G to C, B and M using ResNet50.}\label{ablation_tau}
\end{table}

\begin{table}[t]
\centering
\begin{tabular}[t]{c|cccc}
\hline
$q$ & 5 &10& 20&30\\
\hline
Mean mIoU& 43.94&43.94&43.91&43.94\\
\hline
\end{tabular}
\caption{Sensitivity to parameter $q$ on task G to C, B and M using ResNet50.}\label{ablation_q}
\end{table}

\begin{table}[t]
\centering
\begin{tabular}[t]{c|c}
\hline
Method & Mean mIoU\\
\hline
SSDP (w/o Aug)& 35.83 \\
SSDP (w/ IA)& 43.94\\
SSDP (w/ IA\&GA)& 44.20\\
\hline
\end{tabular}
\caption{Results of different types of augmentations for SSDP over the task G to C, B and M using ResNet50.}\label{ablation_aug_ssdp}
\end{table}

\noindent
\textbf{Comparison of different types of augmentation in Self-supervised Source Domain Projection (SSDP).} We research the impact of different types of augmentations used in our SSDP have on the overall results. Specifically, we investigate the Image Augmentation (IA), which is consist of contrast, brightness, hue, saturation and noise, and Geometric Augmentation (GA) which is composed of crop, scale and flip. As shown in Table~\ref{ablation_aug_ssdp}, the first row is the result of which we train our SSDP without any data augmentation, in this scene our SSDP can be regarded as a standard Auto-Encoder. In the second row, we use IA strategy in SSDP and improve mean mIoU from 35.83 to 43.94.  As for the last row,  we first apply the GA strategy to the original data and further utilized IA, and our SSDP is trained to project the data both used GA and IA to the data only used GA. Our SSDP trained with both IA and GA strategies achieves the best performance, which implies various data augmentation strategies may enhance the ability of our SSDP to project the unseen target domain data to the source domain.

\begin{table}[t!]
\centering
\begin{tabular}[t!]{c|cc}
\hline
Method & InfoNCE & Ours\\
\hline
Mean mIoU& 42.21 & 43.94 \\
\hline
\end{tabular}
\caption{Results of different instance-to-class contrastive losses for task G to C, B and M using ResNet50.}\label{instancetoclassloss}
\end{table}
\noindent
\textbf{Comparison of different choices of instance-to-class contrastive learning.} 
We compare the performance of InfoNCE loss \cite{infonce} and the margin triplet loss as our instance-to-class contrastive loss. The results in Table~\ref{instancetoclassloss} show that the margin triplet loss in Eq.~\eqref{loss4} achieves better performance than the InfoNCE loss in our framework.

\noindent
\textbf{Comparison of results on each semantic class.} 
We compare each class performance of Baseline which is trained by cross entropy loss on the source domain, IBN-Net \cite{ibn}, ISW \cite{isw} and our method on the task GTAV to Cityscapes with ResNet50. As shown in Table~\ref{class_iou}, DPCL achieves the best performance in 15 classes among 19 classes, such as Road, TrafficSign (TS), Rider, Bus, Motor and Bike. Especially for the class Bike, our method improves the IoU from 12.20 to 45.10. As for class Vegetation (Veg.), Terrain and Car, DPCL achieves comparable results with compared methods. Additionally, while the compared methods struggle to recognize the class Train, our DPCL shows an improved IoU for this class.

\begin{table*}[htbp]
\centering
\setlength{\tabcolsep}{1.4pt}
\scalebox{0.9}{
\begin{tabular}[htbp]{c|ccccccccccccccccccc|c}
\hline
Method&Road&SW&Build&Wall&Fence&Pole&TL&TS&Veg.&Terrain&Sky&Person&Rider&Car&Truck&Bus&Train&Motor&Bike&mIoU\\
\hline
Baseline&45.10&23.13&56.77&16.63&16.30&23.90&30.00&13.37&80.87&24.27&38.93&58.17&7.17&61.03&20.03&17.40&\underline{1.17}&8.50&7.33&28.95\\

IBN-Net&51.27&24.07&59.73&14.07&\textbf{25.93}&23.03&30.90&\underline{15.73}&\underline{85.03}&\textbf{40.63}&67.83&60.63&4.93&76.67&23.67&16.27&0.83&11.87&10.07&33.85\\

ISW&\underline{60.47}&\underline{25.53}&\underline{65.40}&\underline{21.57}&\underline{23.73}&\underline{25.77}&\underline{33.33}&15.47&\textbf{85.43}&\underline{38.50}&\underline{70.30}&\underline{61.87}&\underline{9.30}&\textbf{82.73}&\underline{25.43}&\underline{21.07}&0.03&\underline{16.80}&\underline{12.20}&\underline{36.58}\\
DPCL  &\textbf{77.10}&\textbf{33.70}&\textbf{72.00}&\textbf{27.90}&21.90&\textbf{33.70}&\textbf{42.90}&\textbf{28.40}&84.40&34.50&\textbf{80.30}&\textbf{63.30}&\textbf{22.90}&\underline{82.40}&\textbf{28.30}&\textbf{35.80}&\textbf{7.90}&\textbf{30.00}&\textbf{45.10}&\textbf{44.87}\\
\hline

\end{tabular}}
\caption{Results of each semantic class on the domain generalization semantic segmentation task GTAV to Cityscapes using ResNet50. The best and
second best results are bolded and underlined respectively.}\label{class_iou}
\end{table*}

\begin{figure*}[t]   
	\centering
	\includegraphics[width=\linewidth,scale=0.50]{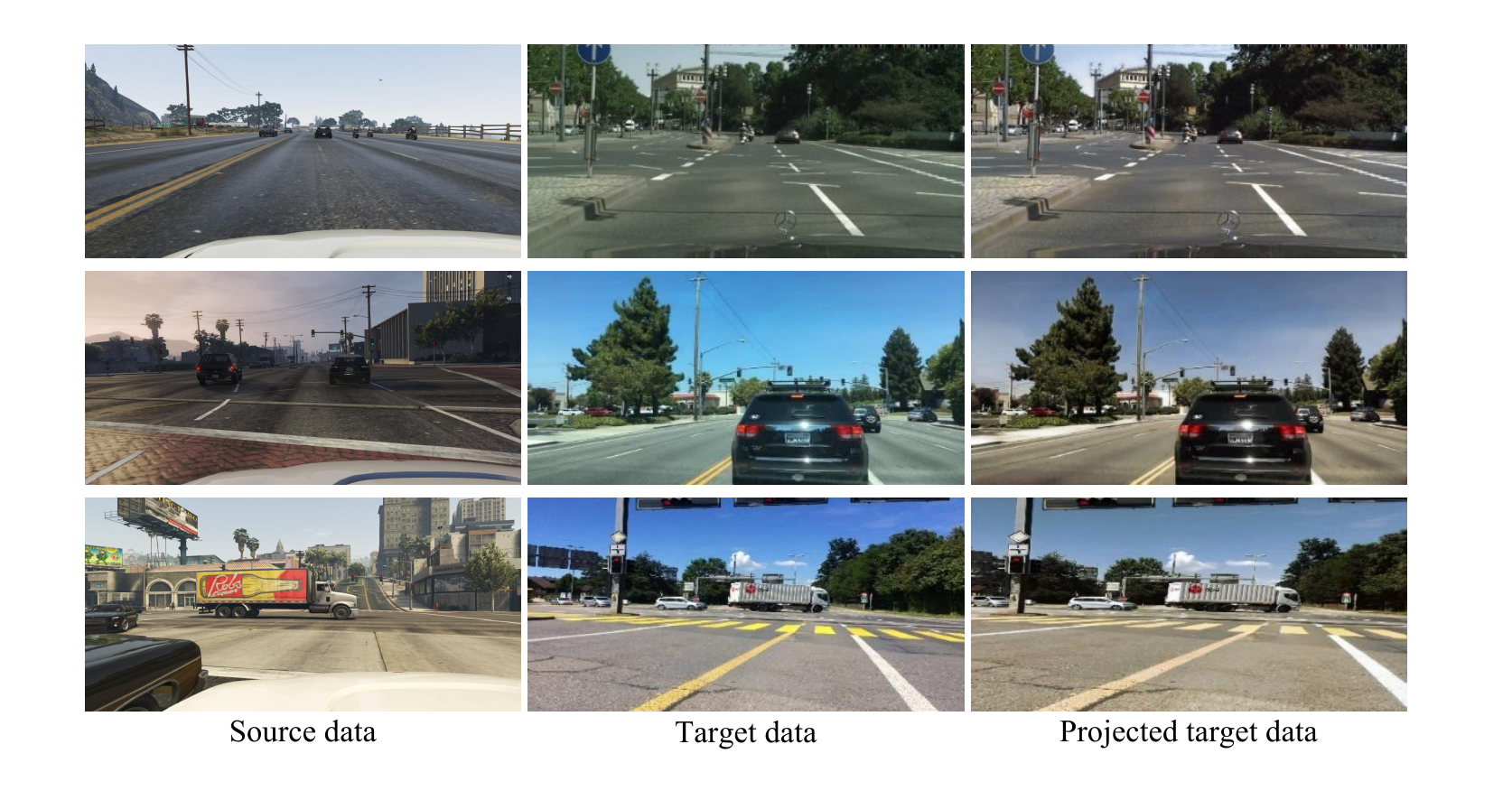}
	\caption{Visualization of our projected target data on the task from GTAV \cite{gtav} to Cityscapes \cite{cityscapes}, BDD \cite{bdd100k} and Mapillary \cite{mapillary}.}
	\label{projected_task1}
\end{figure*}

\begin{figure*}[h]   
	\centering
	\includegraphics[width=\linewidth,scale=0.50]{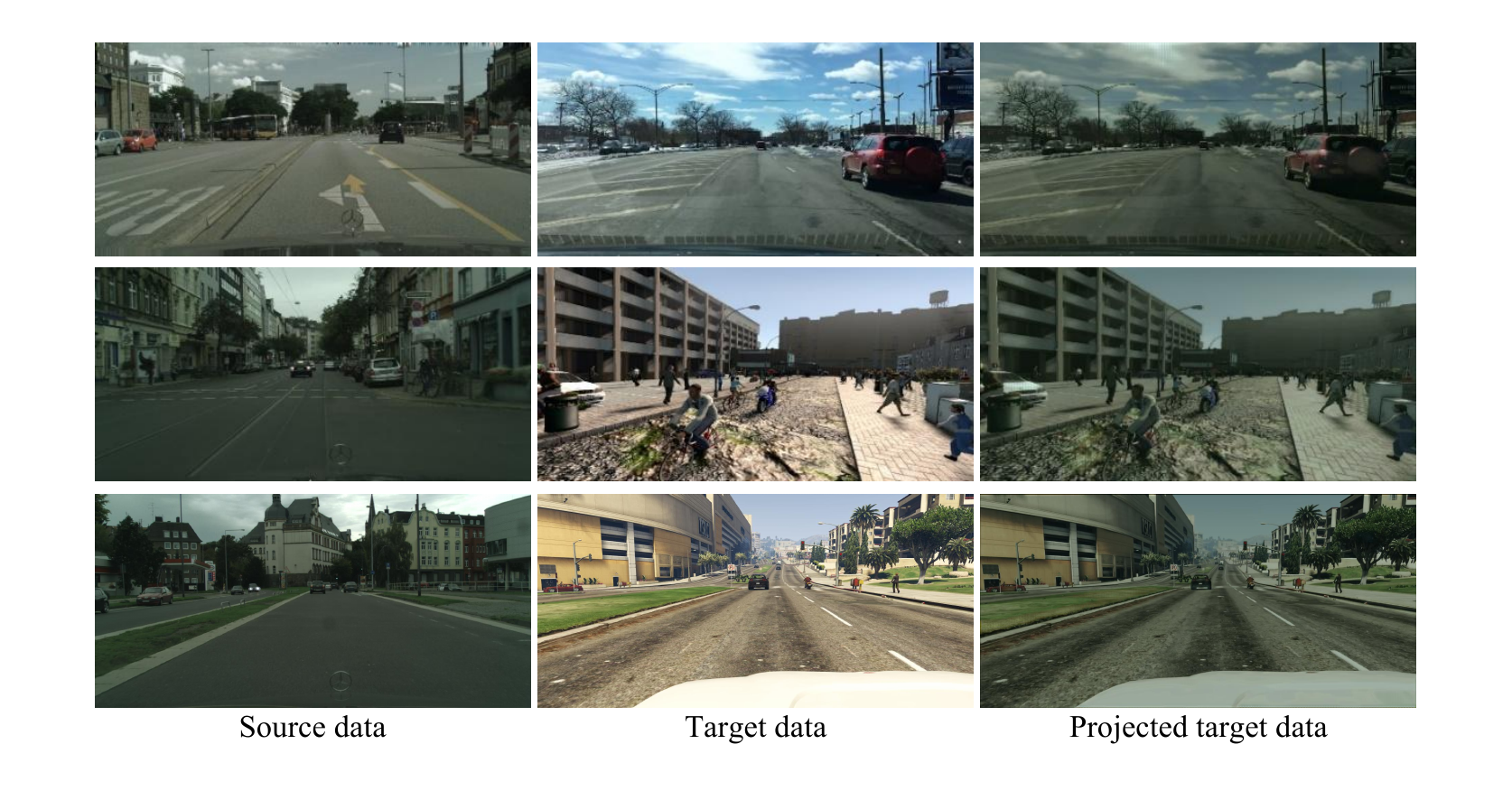}
	\caption{Visualization of our projected target data on the task from Cityscapes \cite{cityscapes} to BDD \cite{bdd100k}, Synthia \cite{synthia} and GTAV \cite{gtav}.}
	\label{projected_task2}
\end{figure*}
\begin{figure*}[h]   
	\centering
	\includegraphics[width=\linewidth,scale=0.70]{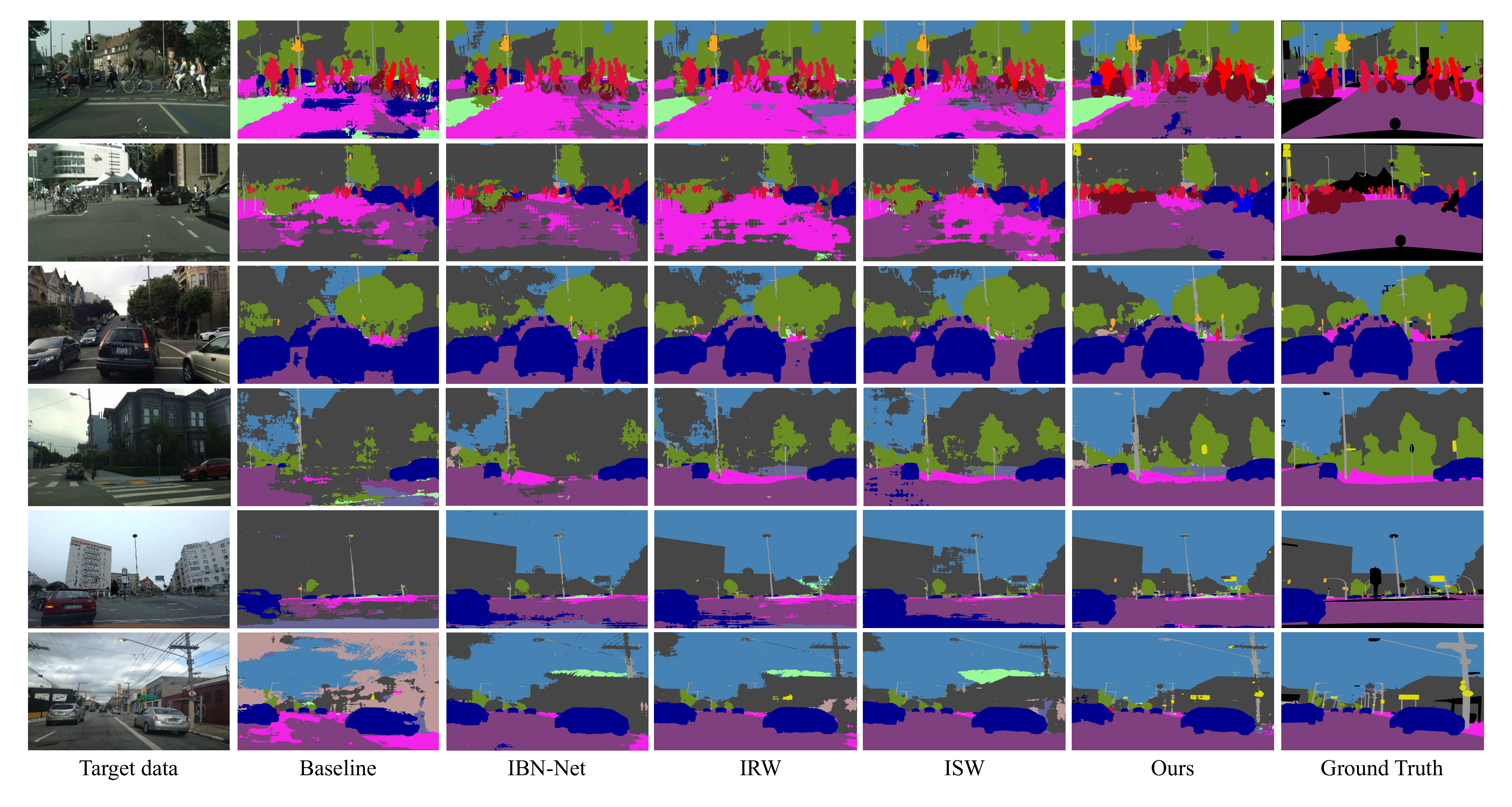}
	\caption{Visual comparison of different domain generalization semantic segmentation methods using ResNet50, trained on GTAV (G) \cite{gtav} and tested on unseen target domains of Cityscapes~\cite{cityscapes}, BDD ~\cite{bdd100k} and Mapillary~\cite{mapillary}.}
	\label{task_1}
\end{figure*}
\section{D. More qualitative results}
In this section, we show the projected images by our Self-supervised Source Domain Projection (SSDP) and more segmentation results compared with the other methods. As shown in Fig.~\ref{projected_task1} and Fig.~\ref{projected_task2}, we visualize the target data and projected data in the last two columns. The projected target data has a closer style to the source data in both two tasks, which illustrates that our SSDP can generate projected data closer to the source domain. 

Figure~\ref{task_1} shows more qualitative results for the task GTAV to Cityscapes, BDD and Mapillary using ResNet50. Our method is superior than the compared methods IBN \cite{ibn} , IRW \cite{isw} and ISW \cite{isw}.
\section*{Acknowledgements}
This work was supported by National Key R\&D Program 2021YFA1003002, NSFC (12125104, U20B2075, 11971373, 61721002, U1811461), and the Fundamental Research Funds for the Central Universities.
\bibliography{aaai23}

\end{document}